\documentclass{article}
\usepackage{amsmath,graphicx,mlspconf}

\copyrightnotice{}

\copyrightnotice{}

\copyrightnotice{}

\copyrightnotice{}

\toappear{}

\usepackage{color}
\definecolor{darkgreen}{rgb}{0,0.6,0.2}

\title{UNCERTAINTY MODELING AND INTERPRETABILITY IN CONVOLUTIONAL NEURAL NETWORKS FOR POLYP SEGMENTATION}

\name{Kristoffer Wickstr\o m*, Michael Kampffmeyer, Robert Jenssen \thanks{This work is partially funded by the Norwegian Research Council FRIPRO grant no. 239844 on developing the \emph{Next Generation Learning Machines}.}\thanks{*Corresponding author: kwi030@uit.no}}
\address{UiT The Arctic University of Norway \\
        Dept. of Physics and Technology \\
        UiT Machine Learning Group \\
        }

\begin{document}
%\ninept
%

\maketitle
\begin{abstract}
Convolutional Neural Networks (CNNs) are propelling advances in a range of different computer vision tasks such as object detection and object segmentation. Their success has motivated research in applications of such models for medical image analysis. If CNN-based models are to be helpful in a medical context, they need to be precise, interpretable, and uncertainty in predictions must be well understood. In this paper, we develop and evaluate recent advances in uncertainty estimation and model interpretability in the context of semantic segmentation of polyps from colonoscopy images. We evaluate and enhance several architectures of Fully Convolutional Networks (FCNs) for semantic segmentation of colorectal polyps and provide a comparison between these models. Our highest performing model achieves a 76.06\% mean IOU accuracy on the EndoScene dataset, a considerable improvement over the previous state-of-the-art.
\end{abstract}
\begin{keywords}
Polyp Segmentation, Deep Learning, Fully Convolutional Networks, Uncertainty Modeling, CNN interpretability
\end{keywords}
\section{Introduction}

Colon cancer prevention is currently primarily done with the help of regular colonoscopy screenings. However, depending on the size and type, roughly $8-37\%$ of polyps are missed during the process~\cite{Missrate}. Missed polyps can have fatal consequences, as they are potential precursors to colon cancer, which causes the third most cancer deaths globally~\cite{DBLP:journals/corr/VazquezBSFLRDC16}. Hence, increasing the detection rate of polyps is an important topic of research. Towards this end, automated detection procedures have been proposed ~\cite{journals/cmig/BernalSFGRV15-clinic,Bernal20142}, which have the advantage of not being influenced by factors such as the fatigue of medical personnel towards the end of long operations. However, they present additional novel challenges. For instance, to enable effective use of such methods, medical staff should be able to understand why the model believes that a specific region contains a polyp (interpretability) and there should be some notion of uncertainty in predictions. 

The last couple of years have seen several works on automatic procedures based on Deep Convolutional Neural Networks~(CNN) for health domain tasks such as interstitial lung disease classification~\cite{7064414}, cell detection~\cite{xue2017cell}, estimation of cardiothoracic ratio~\cite{dong2018unsupervised}, and polyp detection~\cite{7545996}. CNNs have improved the state of the art in a number of computer vision tasks such as image classification~\cite{NIPS2012}, object detection~\cite{DBLP:journals/corr/RenHG015} and semantic segmentation~\cite{DBLP:journals/corr/ShelhamerLD16,DBLP:journals/corr/ChenPKMY14}. Recently, CNNs have shown promising performance for the task of polyp segmentation~\cite{DBLP:journals/corr/VazquezBSFLRDC16}. However, despite the promising results obtained on polyp segmentation, model interpretability and uncertainty quantification have been lacking, and recent advances in deep learning have not been incorporated~\cite{DBLP:journals/corr/VazquezBSFLRDC16}.

In this paper, we enhance and evaluate two recent CNN architectures for pixel-to-pixel based segmentation of colorectal polyps, referred to as Fully Convolutional Networks (FCNs). Furthermore, we incorporate and develop recent advances in the field of deep learning for semantic segmentation of colorectal polyps in order to model uncertainty and increase model interpretability leading to novel uncertainty maps~\cite{DBLP:journals/corr/KendallBC15, kampffmeyer2016semantic} in a polyp segmentation context as well as the visualization of descriptive regions in the input image using Guided Backpropagation~\cite{DBLP:journals/corr/SpringenbergDBR14}. To the author's knowledge, these techniques have not been previously explored in the field of semantic segmentation of colorectal polyps.

\section{Enhanced Fully Convolutional Networks for Polyp Segmentation}

We choose two architectures for the task of polyp segmentation, namely the Fully Convolutional Network 8 (FCN-8)~\cite{DBLP:journals/corr/ShelhamerLD16} and the more recent SegNet~\cite{DBLP:journals/corr/BadrinarayananK15}. Previous use of FCNs for polyp segmentation have shown promising results, and we hypothesized that the inclusion of recent advances in deep learning would improve these results further. SegNet has been shown to achieve comparative results to FCNs in some application domains but is a less memory intensive approach with fewer parameters to optimize. 

The FCN is a CNN architecture particularly well suited to tackle per pixel prediction problems like semantic segmentation. FCNs employ an encoder-decoder architecture and are capable of end-to-end learning. The encoder extracts useful features from an image and maps it to a low-resolution representation. The decoder is tasked with mapping the low-resolution representation back into the same resolution as the input image. Upsampling in FCNs is performed using either bi-linear interpolation or by transposed (or fractional strided) convolutions, where the convolution filters are learned as part of the optimization procedure. Learned upsample filters add additional parameters to the network architecture, but tend to provide better overall results~\cite{DBLP:journals/corr/ShelhamerLD16}. Upsampling can further be improved by including skip connections, which combine coarse level semantic information with higher resolution segmentation from previous network layers. Due to the lack of fully connected layers, inference can be performed on images of arbitrary size.

The SegNet architecture builds on the general idea of FCNs but proposes a novel approach to upsampling. Instead of learning the upsampling, SegNet utilizes the pooling indices from the encoder to upsample activations in the decoder, thereby producing sparse feature maps. These sparse representations are then processed by additional convolutional layers to produce dense activations and predictions. The advantage of SegNet compared to FCN is the reduction in learnable parameters, only 29.5 vs. 134.5 million, as the upsampling filters in FCNs tend to be large.
%The advantage of SegNet is that the number of parameters that have to be learned is considerably lower than the FCN, only 29.5 vs. 134.5 million since the upsampling filters in FCNs tend to be large.

\begin{figure}
    \centering
    \centerline{\includegraphics[width=8.5cm]{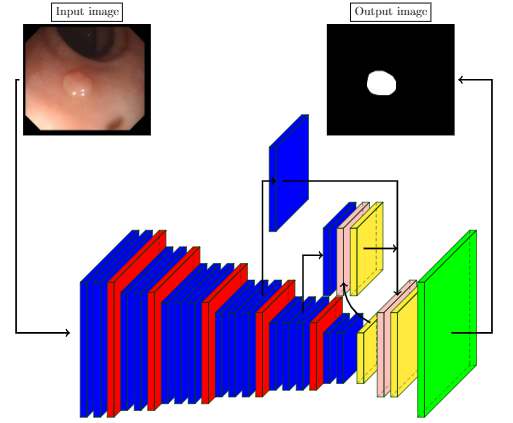}}
    \caption{An illustration of the Enhanced Fully Convolutional Network-8. Color codes description: Blue - Convolution (3x3), Batch Normalization and ReLU; Yellow - Upsampling; Pink - Summing; Red - Pooling (2x2); Green - Soft-max. Dropout was included according to~\cite{DBLP:journals/corr/SimonyanZ14a}.}
    \label{fig:fcn8}
\end{figure}

\begin{figure}
    \centering
    \centerline{\includegraphics[width=8.5cm]{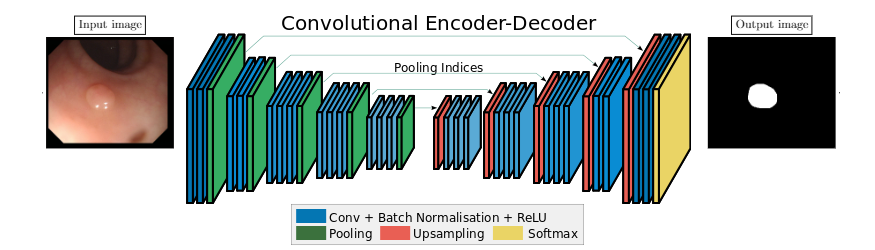}}
    \caption{Depiction of the standard SegNet architecture obtained from~\cite{DBLP:journals/corr/BadrinarayananK15}. Our implementation include Dropout in the three central encoders and decoders for regularization and to enable uncertainty estimation.}
    \label{fig:segnet}
\end{figure}

For our experiments, we propose an Enhanced FCN-8 (EFCN-8) architecture which leverages recent development in the deep learning field. The architecture is depicted in Figure~\ref{fig:fcn8}. The second architecture that we propose is an Enhanced variant of SegNet (ESegNet), depicted in Figure~\ref{fig:segnet}. To improve overall performance, we propose to include several recent advances in deep learning which were not present in the original architectures. For the FCN-8, we make use of Batch Normalization~\cite{DBLP:journals/corr/IoffeS15} after each layer. Batch Normalization is a procedure that normalizes the output of each layer, allowing for a larger learning rate that accelerate the training procedure. For ESegNet, we include Dropout~\cite{JMLR:v15:srivastava14a} after the three central encoders and decoders inspired by~\cite{DBLP:journals/corr/KendallBC15}. Dropout is a regularization technique that randomly set units in a layer to zero and can be interpreted as an ensemble of several networks. Including Dropout serves two purposes. It regularizes the model which encourage better generalization capabilities and, as we will see in Section~3, enable estimation of uncertainty in the model's prediction. The encoder of both models corresponds to the network VGG16~\cite{DBLP:journals/corr/SimonyanZ14a}, which allows us to initialize the encoder with pretrained weights from a VGG16 model that was previously trained on the ImageNet dataset, an approach referred to as transfer learning. Utilizing pretrained weights was incorporated in the original architectures, but not included in recent work on segmentation of colorectal polyp using FCNs~\cite{DBLP:journals/corr/VazquezBSFLRDC16}.

\section{Uncertainty and Interpretability In Fully Convolutional Networks for polyp Segmentation}

Despite their success on a number of different tasks, CNNs are not without their flaws. Most CNNs are unable to provide any notion of uncertainty in their prediction and determining what features in the input affect its prediction is challenging as a result of the complexity of the model. Such limitations have not stopped CNNs from being applied on many computer vision tasks, yet they become especially apparent when developing CNNs for medical applications. Most physicians will be reluctant to make a diagnosis based on a single segmentation map with no notion of uncertainty and indication as to what features were the basis of the prediction. This section will describe two recently proposed methods which address the limitations just outlined.

\subsection{Uncertainty}

Modeling uncertainty is crucial for designing trustworthy automatic procedures, yet CNNs have no natural way of providing such uncertainties to accompany its prediction. In contrast, Bayesian models provide a framework which naturally includes uncertainty by modeling posterior distribution for the quantities in question. Given a dataset $\mathbf{X}=\{\mathbf{x}_1,...,\mathbf{x}_N\}$ with labels $\mathbf{Y}=\{\mathbf{y}_1,...,\mathbf{y}_N\}$, the predictive distribution of a Bayesian neural network can be modeled as:

\begin{equation}\label{eq:bayes predictive}
    p(\mathbf{y}|\mathbf{x},\mathbf{X},\mathbf{Y})=\int p(\mathbf{y}|\mathbf{x},\mathbf{W})p(\mathbf{W|\mathbf{x},\mathbf{X},\mathbf{Y}})d\mathbf{W}
\end{equation}

where $\mathbf{W}$ refers to the weights of the model, $p(\mathbf{y}|\mathbf{x},\mathbf{W})$ is the softmax function applied to the output of the model, denoted by $f^\mathbf{W}(\mathbf{x})$, and $p(\mathbf{W|\mathbf{x},\mathbf{X},\mathbf{Y}})$ is the posterior over the weights which capture the set of plausible models parameters for the given data. Obtaining $p(\mathbf{y}|\mathbf{x},\mathbf{W})$ only requires a forward pass of the network, but the inability to evaluate the posterior of the weights analytically makes Bayesian neural networks computationally unfeasible. To sidestep the problematic posterior of the weights,~\cite{Gal} proposed to incorporate Dropout~\cite{JMLR:v15:srivastava14a} as a method for sampling sets of weights from the trained network to approximate the posterior of the weights. The predictive distribution in Equation~\ref{eq:bayes predictive} can then be approximated using Monte Carlo integration as follows:

\begin{equation}\label{eq:approx predictive}
    p(\mathbf{y}|\mathbf{x},\mathbf{X},\mathbf{Y})\approx\frac{1}{T}\sum\limits_{t=1}^T \text{Softmax}(f^{\widehat{\mathbf{W}}_t}(\mathbf{x}))
\end{equation}

where $T$ is the number of sampled sets of weights and $\widehat{\mathbf{W}}_t$ is a set of sampled weights. In practice, the predictive distribution from Equation~\ref{eq:approx predictive} can be estimated by running $T$ forward passes of a model with Dropout applied to produce $T$ predictions, which in turn can be used to estimate the uncertainty associated with the sample in question. The authors of~\cite{Gal} refer to this method of sampling from the posterior of the predictive distribution as Monte Carlo Dropout.

\subsection{Interpretability}

Another desirable property that CNNs lack is interpretability, i.e. being able to determine what features induces the network to produce a particular prediction. However, several recent works have proposed different methods to increase network interpretability~\cite{DBLP:journals/corr/ZeilerF13, bach-plos15}. In this paper, we evaluate and develop the Guided Backpropagation~\cite{DBLP:journals/corr/SpringenbergDBR14} technique for FCNs on the task of semantic segmentation of colorectal polyp in order to assess which pixels in the input image the network deems important for identifying polyps. We choose Guided Backpropagation as it is known to produce clear visualizations of salient input pixels and is more straightforward to employ compared to other methods.

The central idea of Guided Backpropagation is the interpretation of the gradients of the network with respect to an input image. In~\cite{DBLP:journals/corr/SimonyanVZ13} they noted that, for a given image, the magnitude of the gradients indicate which pixels in the input image need to be changed the least to affect the prediction the most. By utilizing backpropagation, they obtained the gradients corresponding to each pixel in the input such that they could visualize what features the network considers essential. In~\cite{DBLP:journals/corr/SpringenbergDBR14} they argued that positive gradients with a large magnitude indicate pixels of high importance while negative gradients with a large magnitude indicate pixels which the networks want to suppress which, if included in the visualization of important pixels, can result in noisy images. To avoid this, Guided Backpropagation alters the backward pass of a neural network such that negative gradients are set to zero in each layer, thus allowing only positive gradients to flow backward through the network and highlighting pixels which the system finds important.

\section{Results}

In this section, we present quantitative and qualitative results on semantic segmentation of colorectal polyps for both architectures, along with details regarding the training of the two models. We also present the results of using Monte Carlo Dropout to model the uncertainty associated with the predictions and the results of using Guided Backpropagation to visualize which pixels are considered important.

\subsection{Training Approach}

We evaluate our methods on the EndoScene~\cite{DBLP:journals/corr/VazquezBSFLRDC16} dataset for semantic segmentation of colorectal polyps, which consists of 912 RGB images obtained from colonoscopies from 36 patients. For each of the input images comes a corresponding annotated image provided by physicians, where pixels belonging to a polyp are marked in white and pixels belonging to the colon are marked in black. The first row of Figure~\ref{fig:qualitativ results2} and~\ref{fig:qualitativ results} display examples from the dataset. We consider the two-class problem, where the task is to classify each pixel as polyp or part of the colon (background class). Following the approach of~\cite{DBLP:journals/corr/VazquezBSFLRDC16} we separate the dataset into training/validation/test set, where the training set consists of 20 patients and 547 images, the validation set consists of 8 patients and 183 images, and the test set consists of 8 patients and 182 images. All RGB input image are normalized to range $[0,1]$.

For performance evaluation, we calculate Intersection over Union (IoU) and global accuracy (per-pixel accuracy) on the test set. For a given class $c$, prediction $\hat{y}_i$ and ground truth $y_i$, the IoU is defined as

\begin{equation}
    \text{IoU}(c) = \frac{\sum_i (\hat{y}_i==c \land y_i==c )}{\sum_i (\hat{y}_i==c \lor y_i==c )}
\end{equation}

where $\land$ is the logical $and$ operation and $\lor$ is the logical $or$ operation.

We initialize the decoder weights of both EFCN-8 and ESegNet using HeNormal initialization~\cite{DBLP:journals/corr/HeZR015} and employ pre-trained weights for the encoders, as mentioned previously. All models were trained using ADAM~\cite{DBLP:journals/corr/KingmaB14} with a batch size of 10 and cross-entropy loss~\cite{DBLP:journals/corr/ShelhamerLD16}. We use the validation set to apply early stopping and monitor polyp IoU score with a patience of 30. Class balancing was not applied as it gave no significant improvement and no weight decay was used.

Data augmentation was applied according to best practices to increase the number of training images artificially. We utilize a dynamic augmentation scheme that applies cropping, rotation, zoom, and shearing. During training, we crop images into 224x224 patches randomly chosen from the center or one of the corners, following the example of~\cite{NIPS2012}. We apply random rotation between -90 and 90 degrees, random zoom from 0.8-1.2 and random shearing from 0-0.4.

\subsection{Quantitative and Qualitative Results}

\begin{figure}[htb!]

\begin{minipage}[b]{.47\linewidth}
  \centering
  \centerline{\includegraphics[width=4.0cm]{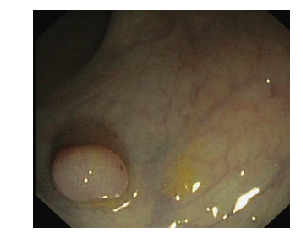}}
%  \vspace{1.5cm}
  \centerline{(a) input image}\medskip
\end{minipage}
\hfill
\begin{minipage}[b]{0.47\linewidth}
  \centering
  \centerline{\includegraphics[width=4.0cm]{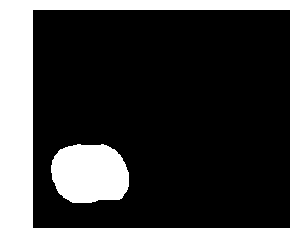}}
%  \vspace{1.5cm}
  \centerline{(b) Ground truth}\medskip
\end{minipage}
\begin{minipage}[b]{.47\linewidth}
  \centering
  \centerline{\includegraphics[width=4.0cm]{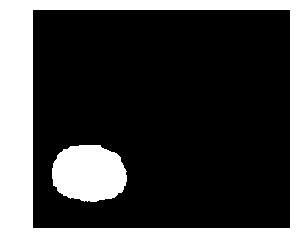}}
%  \vspace{1.5cm}
  \centerline{(c) EFCN-8 prediction}\medskip
\end{minipage}
\hfill
\begin{minipage}[b]{0.47\linewidth}
  \centering
  \centerline{\includegraphics[width=4.0cm]{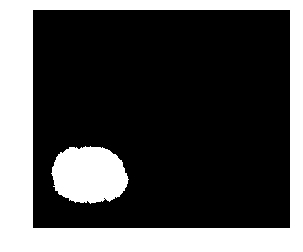}}
%  \vspace{1.5cm}
  \centerline{(d) ESegNet prediction}\medskip
\end{minipage}
\begin{minipage}[b]{.47\linewidth}
  \centering
  \centerline{\includegraphics[width=4.0cm]{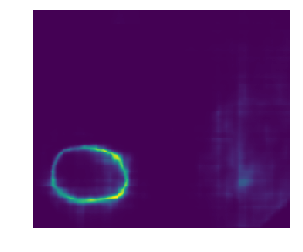}}
%  \vspace{1.5cm}
  \centerline{(e) EFCN-8 uncertainty}\medskip
\end{minipage}
\hfill
\begin{minipage}[b]{0.47\linewidth}
  \centering
  \centerline{\includegraphics[width=4.0cm]{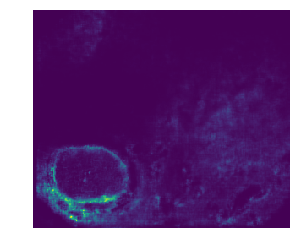}}
%  \vspace{1.5cm}
  \centerline{(f) ESegNet uncertainty}\medskip
\end{minipage}
\begin{minipage}[b]{.47\linewidth}
  \centering
  \centerline{\includegraphics[width=4.0cm]{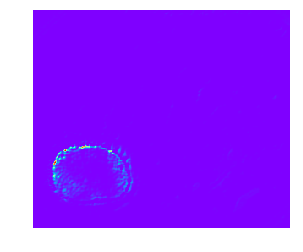}}
%  \vspace{1.5cm}
  \centerline{(g) EFCN-8 interpretability}\medskip
\end{minipage}
\hfill
\begin{minipage}[b]{0.47\linewidth}
  \centering
  \centerline{\includegraphics[width=4.0cm]{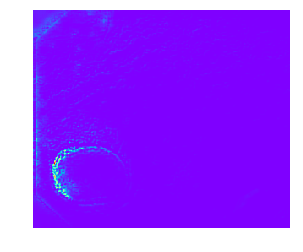}}
%  \vspace{1.5cm}
  \centerline{(h) ESegNet interpretability}\medskip
\end{minipage}
\caption{Qualitative results on the Endoscene test set. The first row, from left to right, displays input image and its corresponding ground truth. The second row displays the prediction of both models. The third row displays the uncertainty associated with the prediction for both models and the fourth row displays which features are highlighted as important for both models.}
\label{fig:qualitativ results2}
\end{figure}

In this section, we present the results for both architectures. Table~\ref{tab:results} presents the quantitative results for our EFCN-8 and ESegNet on the test set along with previous results on this dataset obtained for a vanilla FCN-8~\cite{DBLP:journals/corr/VazquezBSFLRDC16} and a previously state-of-the-art methods based on non-deep learning methods~\cite{Bernal20142}. Row two of Figure~\ref{fig:qualitativ results2} and~\ref{fig:qualitativ results} displays predictions from both models based on samples from the test set.

\begin{table}[htb]
    \centering
    \resizebox{\columnwidth}{!}{%
    \begin{tabular}{l|r|c|c||c|c}
        \textbf{Model} & \# \textbf{P(M)}
         & \textbf{IoU B} & \textbf{IoU P}
         & \textbf{IoU M} & \textbf{Acc M}\\
        \hline
        SDEM~\cite{Bernal20142} & $<<1$  & 0.739 & 0.221 & 0.480 & 0.756 \\
        \hline
        FCN-8~\cite{DBLP:journals/corr/VazquezBSFLRDC16} & 134.5 & \textbf{0.946} & 0.509 & 0.728 & \textbf{0.949} \\
        \hline
        ESegNet & 29.5  & 0.933 & 0.522 & 0.728 & 0.935 \\  
        \hline
        %\multicolumn{6}{l}{\textbf{Model}} \\ \hline
        EFCN-8 & 134.5 & \textbf{0.946} & \textbf{0.587} & \textbf{0.767} & \textbf{0.949} \\
    \end{tabular}}
    \caption{Results for the two-class problem of the EndoScene dataset. Abbreviations are: \# P(M) = number of parameters in millions, IoU (Background, Polyp and Mean) and Accuracy Mean.}
    \label{tab:results}
\end{table}

\subsection{Uncertainty and Interpretability Results}

\begin{figure}[htb!]

\begin{minipage}[b]{.47\linewidth}
  \centering
  \centerline{\includegraphics[width=4.0cm]{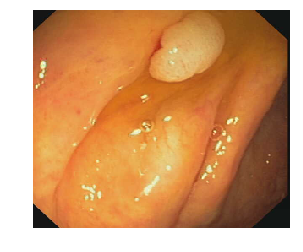}}
%  \vspace{1.5cm}
  \centerline{(a) input image}\medskip
\end{minipage}
\hfill
\begin{minipage}[b]{0.47\linewidth}
  \centering
  \centerline{\includegraphics[width=4.0cm]{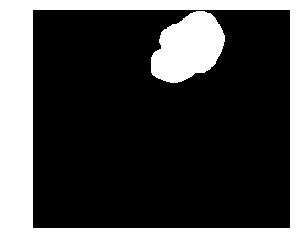}}
%  \vspace{1.5cm}
  \centerline{(b) Ground truth}\medskip
\end{minipage}
\begin{minipage}[b]{.47\linewidth}
  \centering
  \centerline{\includegraphics[width=4.0cm]{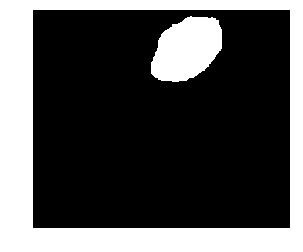}}
%  \vspace{1.5cm}
  \centerline{(c) EFCN-8 prediction}\medskip
\end{minipage}
\hfill
\begin{minipage}[b]{0.47\linewidth}
  \centering
  \centerline{\includegraphics[width=4.0cm]{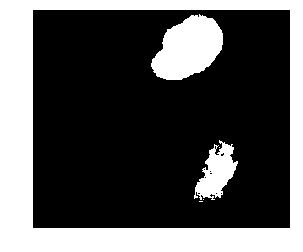}}
%  \vspace{1.5cm}
  \centerline{(d) ESegNet prediction}\medskip
\end{minipage}
\begin{minipage}[b]{.47\linewidth}
  \centering
  \centerline{\includegraphics[width=4.0cm]{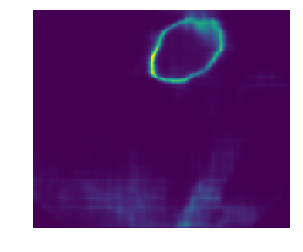}}
%  \vspace{1.5cm}
  \centerline{(e) EFCN-8 uncertainty}\medskip
\end{minipage}
\hfill
\begin{minipage}[b]{0.47\linewidth}
  \centering
  \centerline{\includegraphics[width=4.0cm]{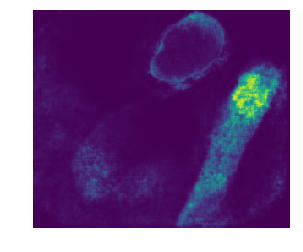}}
%  \vspace{1.5cm}
  \centerline{(f) ESegNet uncertainty}\medskip
\end{minipage}
\begin{minipage}[b]{.47\linewidth}
  \centering
  \centerline{\includegraphics[width=4.0cm]{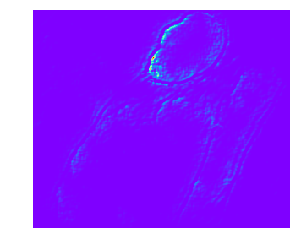}}
%  \vspace{1.5cm}
  \centerline{(g) EFCN-8 interpretability}\medskip
\end{minipage}
\hfill
\begin{minipage}[b]{0.47\linewidth}
  \centering
  \centerline{\includegraphics[width=4.0cm]{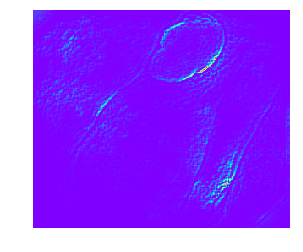}}
%  \vspace{1.5cm}
  \centerline{(h) ESegNet interpretability}\medskip
\end{minipage}
\caption{Qualitative results on the Endoscene test set. The first row, from left to right, displays input image and its corresponding ground truth. The second row displays the prediction of both models. The third row displays the uncertainty associated with the prediction for both models and the fourth row displays which features are highlighted as important for both models.}
\label{fig:qualitativ results}
\end{figure}

The third row of Figure~\ref{fig:qualitativ results2} and~\ref{fig:qualitativ results} show the estimated standard deviation of each pixel in the predictions of both models. Using Monte Carlo Dropout, we obtain ten predictions which we used to estimate the standard deviation of each pixel. Row four of Figure~\ref{fig:qualitativ results2} and~\ref{fig:qualitativ results} displays the results of using Guided Backpropagation to highlight what pixels in the input image both models consider important to their prediction.

\section{Discussion}

From Table~\ref{tab:results} it is evident that the deep learning based models provide predictions of higher precision compared to previous methods based on traditional machine learning techniques. Also, the difference in performance between~\cite{DBLP:journals/corr/VazquezBSFLRDC16} FCN-8 and our EFCN-8 indicated that the inclusion of recent techniques such as transfer learning and Batch Normalization is vital to increase the capabilities of deep models. Furthermore, ESegNet can achieve comparable results to the FCN-8 even though it has far fewer parameters. However, when recent techniques are included, we get a gap in performance between ESegNet and the EFCN-8. This might imply that the increased complexity of the EFCN-8 is beneficial to performance.

For the qualitative results, Figure~\ref{fig:qualitativ results2} displays an example where both models produce a correct prediction while Figure~\ref{fig:qualitativ results} shows an example where both models correctly segment the polyp present in the input image, but ESegNet also predicts a polyp where there is none. We have included the first example to illustrate that both models are capable of producing precise and correct predictions. The second example is included to highlight the difficulty when comparing models. For instance, given the predictions in Figure~\ref{fig:qualitativ results} and no ground truth, which prediction would you trust? We know the EFCN-8 achieved a higher performance on the test set, but that does not necessarily mean that it is correct in this particular case. Without further information, it would be difficult to choose between the two models without consulting a medical expert to asses the images.

However, if we could say something about the uncertainty of the two predictions we might choose the prediction with the lowest uncertainty associated with it. For the uncertainty estimates shown in row three of Figure~\ref{fig:qualitativ results2}, where both models successfully segment the polyp in the image, we see that the only pixels associated with high uncertainty are those around the border of the prediction. Such uncertainties are understandable as even physicians are unable to state precisely where the colon ends and the polyp starts. However, the uncertainty estimates which are shown in row three of Figure~\ref{fig:qualitativ results} tell a different story. Notice that both models exhibit similar uncertainty for the region where they both correctly segment a polyp, but ESegNet also has a large area of uncertainty for pixels associated with a ridge going along the colon. The falsely segmented polyp lies on this ridge of uncertainty, which might indicate that we should be careful about trusting the polyp prediction toward the bottom right of the image.

Another interesting question is, what pixels in the input image is influencing these predictions? In row four of Figure~\ref{fig:qualitativ results2}, we see that pixels associated with the edges of the polyp are highlighted, indicating that the models are leveraging edge information to identify the polyp. Also, notice that the EFCN-8 considers the entire top edge of the polyp while ESegNet only considers the left edge of the polyp, which might imply that the EFCN-8 has obtained a deeper understanding of the shape and form of polyp and can extract more useful information from the input image. In row four of Figure~\ref{fig:qualitativ results} we again see that both models are reacting to edges in the input image. But while the EFCN-8 can correctly identify pixels which belong to an actual polyp, ESegNet is also considering pixels which correspond to ridges of the colon.

Visualizing important pixels and modeling uncertainty is not just important to design automatic procedures which are trustworthy but also, as we have seen, allows for model analysis and model comparison. Of course, the results of such methods are still somewhat open to interpretation, and deep learning would benefit from a more theoretical framework for analyzing models, yet including techniques such as Monte Carlo Dropout and Guided Backpropagation can lead towards a better understanding of CNNs.

\section{Conclusion}

In this paper, we improved and applied two established CNN architectures for pixel-wise segmentation, evaluated their performance on colorectal polyp segmentation and conclude that CNNs can achieve high performance in this context of medical image analysis. We also argued that modeling the uncertainty of the networks output and visualizing descriptive image regions in the input image can increase interpretability and make models based on deep learning more applicable for medical personnel.

The field of deep learning is continually improving and several recent architectures for semantic segmentation such as \cite{DBLP:journals/corr/JegouDVRB16} show promising results. Increasing network interpretability is an active field of research where Relevance Propagation \cite{bach-plos15} is a particularly interesting approach for future experimentation. Post-processing procedures have also shown to improve performance of CNNs and including Conditional Random Fields could improve spatial coherence~\cite{DBLP:journals/corr/ChenPKMY14}.

% -------------------------------------------------------------------------
\bibliographystyle{IEEEbib}
{\footnotesize
\bibliography{Bibliography}}

\begin{thebibliography}{10}

\bibitem{Missrate}
Jeroen~C Van~Rijn, Johannes~B Reitsma, Jaap Stoker, Patrick~M Bossuyt, Sander~J
  Van~Deventer, and Evelien Dekker,
\newblock ``Polyp miss rate determined by tandem colonoscopy: a systematic
  review,''
\newblock {\em The American journal of gastroenterology}, vol. 101, no. 2, pp.
  343, 2006.

\bibitem{DBLP:journals/corr/VazquezBSFLRDC16}
David V{\'a}zquez, Jorge Bernal, F~Javier S{\'a}nchez, Gloria
  Fern{\'a}ndez-Esparrach, Antonio~M L{\'o}pez, Adriana Romero, Michal
  Drozdzal, and Aaron Courville,
\newblock ``A benchmark for endoluminal scene segmentation of colonoscopy
  images,''
\newblock {\em Journal of Healthcare Engineering}, vol. 2017, 2017.

\bibitem{journals/cmig/BernalSFGRV15-clinic}
Jorge Bernal, F~Javier S{\'a}nchez, Gloria Fern{\'a}ndez-Esparrach, Debora Gil,
  Cristina Rodr{\'\i}guez, and Fernando Vilari{\~n}o,
\newblock ``Wm-dova maps for accurate polyp highlighting in colonoscopy:
  Validation vs. saliency maps from physicians,''
\newblock {\em Computerized Medical Imaging and Graphics}, vol. 43, pp.
  99--111, 2015.

\bibitem{Bernal20142}
Jorge Bernal, Joan~Manel N{\'u}{\~n}ez, F~Javier S{\'a}nchez, and Fernando
  Vilari{\~n}o,
\newblock ``Polyp segmentation method in colonoscopy videos by means of
  msa-dova energy maps calculation,''
\newblock in {\em Workshop on Clinical Image-Based Procedures}. Springer, 2014,
  pp. 41--49.

\bibitem{7064414}
Q.~Li, W.~Cai, X.~Wang, Y.~Zhou, D.~D. Feng, and M.~Chen,
\newblock ``Medical image classification with convolutional neural network,''
\newblock in {\em 2014 13th International Conference on Control Automation
  Robotics Vision (ICARCV)}, Dec 2014, pp. 844--848.

\bibitem{xue2017cell}
Yao Xue and Nilanjan Ray,
\newblock ``Cell detection with deep convolutional neural network and
  compressed sensing,''
\newblock {\em arXiv preprint arXiv:1708.03307}, 2017.

\bibitem{dong2018unsupervised}
Nanqing Dong, Michael Kampffmeyer, Xiaodan Liang, Zeya Wang, Wei Dai, and
  Eric~P Xing,
\newblock ``Unsupervised domain adaptation for automatic estimation of
  cardiothoracic ratio,''
\newblock in {\em International Conference on Medical Image Computing and
  Computer Assisted Intervention}. Springer, 2018.

\bibitem{7545996}
E.~Ribeiro, A.~Uhl, and M.~Häfner,
\newblock ``Colonic polyp classification with convolutional neural networks,''
\newblock in {\em 2016 IEEE 29th International Symposium on Computer-Based
  Medical Systems (CBMS)}, 2016, pp. 253--258.

\bibitem{NIPS2012}
Alex Krizhevsky, Ilya Sutskever, and Geoffrey~E Hinton,
\newblock ``Imagenet classification with deep convolutional neural networks,''
\newblock in {\em Advances in Neural Information Processing Systems}, 2012, pp.
  1097--1105.

\bibitem{DBLP:journals/corr/RenHG015}
Shaoqing Ren, Kaiming He, Ross Girshick, and Jian Sun,
\newblock ``Faster r-cnn: Towards real-time object detection with region
  proposal networks,''
\newblock {\em IEEE Transactions on Pattern Analysis and Machine Intelligence},
  vol. 39, no. 6, pp. 1137--1149, 2017.

\bibitem{DBLP:journals/corr/ShelhamerLD16}
Evan Shelhamer, Jonathan Long, and Trevor Darrell,
\newblock ``Fully convolutional networks for semantic segmentation,''
\newblock {\em IEEE Transactions on Pattern Analysis and Machine Intelligence},
  vol. 39, no. 4, pp. 640--651, 2017.

\bibitem{DBLP:journals/corr/ChenPKMY14}
Liang-Chieh Chen, George Papandreou, Iasonas Kokkinos, Kevin Murphy, and Alan~L
  Yuille,
\newblock ``Deeplab: Semantic image segmentation with deep convolutional nets,
  atrous convolution, and fully connected crfs,''
\newblock {\em IEEE Transactions on Pattern Analysis and Machine Intelligence},
  vol. 40, no. 4, pp. 834--848, 2018.

\bibitem{DBLP:journals/corr/KendallBC15}
Alex Kendall, Vijay Badrinarayanan, and Roberto Cipolla,
\newblock ``Bayesian segnet: Model uncertainty in deep convolutional
  encoder-decoder architectures for scene understanding,''
\newblock {\em arXiv preprint arXiv:1511.02680}, 2015.

\bibitem{kampffmeyer2016semantic}
Michael Kampffmeyer, Arnt-Borre Salberg, and Robert Jenssen,
\newblock ``Semantic segmentation of small objects and modeling of uncertainty
  in urban remote sensing images using deep convolutional neural networks,''
\newblock in {\em Proceedings of the IEEE Conference on Computer Vision and
  Pattern Recognition Workshops}, 2016, pp. 1--9.

\bibitem{DBLP:journals/corr/SpringenbergDBR14}
Jost~Tobias Springenberg, Alexey Dosovitskiy, Thomas Brox, and Martin
  Riedmiller,
\newblock ``Striving for simplicity: The all convolutional net,''
\newblock {\em arXiv preprint arXiv:1412.6806}, 2014.

\bibitem{DBLP:journals/corr/BadrinarayananK15}
Vijay Badrinarayanan, Alex Kendall, and Roberto Cipolla,
\newblock ``Segnet: A deep convolutional encoder-decoder architecture for image
  segmentation,''
\newblock {\em IEEE Transactions on Pattern Analysis and Machine Intelligence},
  , no. 12, pp. 2481--2495, 2017.

\bibitem{DBLP:journals/corr/IoffeS15}
Sergey Ioffe and Christian Szegedy,
\newblock ``Batch normalization: Accelerating deep network training by reducing
  internal covariate shift,''
\newblock in {\em International Conference on Machine Learning}, 2015, pp.
  448--456.

\bibitem{JMLR:v15:srivastava14a}
Nitish Srivastava, Geoffrey Hinton, Alex Krizhevsky, Ilya Sutskever, and Ruslan
  Salakhutdinov,
\newblock ``Dropout: A simple way to prevent neural networks from
  overfitting,''
\newblock {\em Journal of Machine Learning Research}, vol. 15, pp. 1929--1958,
  2014.

\bibitem{DBLP:journals/corr/SimonyanZ14a}
Karen Simonyan and Andrew Zisserman,
\newblock ``Very deep convolutional networks for large-scale image
  recognition,''
\newblock {\em International Conference on Learning Representations}, 2015.

\bibitem{Gal}
Yarin Gal and Zoubin Ghahramani,
\newblock ``Dropout as a bayesian approximation: Representing model uncertainty
  in deep learning,''
\newblock in {\em International Conference on Machine Learning}, 2016, pp.
  1050--1059.

\bibitem{DBLP:journals/corr/ZeilerF13}
Matthew~D. Zeiler and Rob Fergus,
\newblock ``Visualizing and understanding convolutional networks,''
\newblock {\em CoRR}, vol. abs/1311.2901, 2013.

\bibitem{bach-plos15}
Sebastian Bach, Alexander Binder, Gr{\'e}goire Montavon, Frederick Klauschen,
  Klaus-Robert M{\"u}ller, and Wojciech Samek,
\newblock ``On pixel-wise explanations for non-linear classifier decisions by
  layer-wise relevance propagation,''
\newblock {\em PloS one}, vol. 10, no. 7, pp. e0130140, 2015.

\bibitem{DBLP:journals/corr/SimonyanVZ13}
Karen Simonyan, Andrea Vedaldi, and Andrew Zisserman,
\newblock ``Deep inside convolutional networks: Visualising image
  classification models and saliency maps,''
\newblock {\em CoRR}, vol. abs/1312.6034, 2013.

\bibitem{DBLP:journals/corr/HeZR015}
Kaiming He, Xiangyu Zhang, Shaoqing Ren, and Jian Sun,
\newblock ``Delving deep into rectifiers: Surpassing human-level performance on
  imagenet classification,''
\newblock {\em CoRR}, vol. abs/1502.01852, 2015.

\bibitem{DBLP:journals/corr/KingmaB14}
Diederik~P. Kingma and Jimmy Ba,
\newblock ``Adam: {A} method for stochastic optimization,''
\newblock {\em CoRR}, vol. abs/1412.6980, 2014.

\bibitem{DBLP:journals/corr/JegouDVRB16}
Simon J{\'e}gou, Michal Drozdzal, David Vazquez, Adriana Romero, and Yoshua
  Bengio,
\newblock ``The one hundred layers tiramisu: Fully convolutional densenets for
  semantic segmentation,''
\newblock in {\em Computer Vision and Pattern Recognition Workshops (CVPRW),
  2017 IEEE Conference on}. IEEE, 2017, pp. 1175--1183.

\end{thebibliography}

\end{document}